\documentclass[runningheads]{llncs}
\usepackage[pdftex]{graphicx}
\usepackage{url} 
\usepackage{float}
\usepackage{hyperref}
\usepackage{amsmath}
\usepackage{tabularx}
\usepackage{hyperref} 
\usepackage{amsfonts}
\usepackage{algorithmic}
\usepackage{algorithm}
\usepackage{fancyhdr}
\usepackage{setspace}
\usepackage{multirow}
\usepackage[graphicx]{realboxes}
\usepackage{svg}
\usepackage{changepage}
\usepackage{booktabs} % for professional tables

\usepackage{xcolor}
\usepackage{microtype}

\begin{document}
%\linespread{0.85}
\title{Little Help Makes a Big Difference: Leveraging Active Learning to Improve Unsupervised Time Series Anomaly Detection}
\titlerunning{Active learning for time series anomaly detection}

\author{Hamza Bodor\inst{1,2} \and
Thai V.\ Hoang\inst{1} \and
Zonghua Zhang\inst{1}}

\authorrunning{H.\ Bodor et al.}

\institute{
% Huawei France Research Center
Paris Research Center, Huawei Technologies France 
\and
École des Ponts ParisTech, France
\\ \email{bodor.hamza@gmail.com}, \email{\{thai.v.hoang, zonghua.zhang\}@huawei.com}
}

\maketitle

\begin{abstract}
Key Performance Indicators (KPI), which are essentially time series data, have been widely used to indicate the performance of telecom networks. 
Based on the given KPIs, a large set of anomaly detection algorithms have been deployed for detecting the unexpected network incidents. 
Generally, unsupervised anomaly detection algorithms gain more popularity than the supervised ones, due to the fact that labeling KPIs is extremely time- and resource-consuming, and error-prone.
However, those unsupervised anomaly detection algorithms often suffer from excessive false alarms, especially in the presence of concept drifts resulting from network re-configurations or maintenance. 
To tackle this challenge and improve the overall performance of unsupervised anomaly detection algorithms, we propose to use active learning to introduce and benefit from the feedback of operators, who can verify the alarms (both false and true ones) and label the corresponding KPIs with reasonable effort. 
Specifically, we develop three query strategies to select the most informative and representative samples to label.
We also develop an efficient method to update the weights of Isolation Forest and optimally adjust the decision threshold, so as to eventually improve the performance of detection model.
The experiments with one public dataset and one proprietary dataset demonstrate that our active learning empowered anomaly detection pipeline could achieve performance gain, in terms of F1-score, more than $50\%$ over the baseline algorithm. 
It also outperforms the existing active learning based methods by approximately $6\%-10\%$, with significantly reduced budget (the ratio of samples to be labeled). 

\end{abstract}

\section{Introduction}
\label{sec:introduction}
Anomaly detection has always been one of the grand challenges, yet an essential capability, in building resilient computer and communication networks. 
Being able to detect anomalies will not only guarantee a timely warning of potential failures in the systems, but also ensure a quick remediation and error correction, which may save a lot of unnecessary expenses. 
As a matter of fact, Key Performances Indicators (KPIs), which are essentially time series data collected over time, have been widely used to assess the health status of networks and services. 
%These KPIs record machine operation status or some %business-related measures over time in order to monitor %these systems. 
Any network failures or unexpected incidents can lead to the significant deviation of KPIs from their normal patterns.
Therefore, KPI based anomaly detection algorithms aim at detecting those deviations with respect to the time series characteristics (e.g., seasonality, trend) or statistical features (e.g., min, max, mean).
For example, the commonly seen anomalies include, but not limited to, spike, dip, continuous bursts, sudden or gradual trend change. 

%KPI anomaly detection aims at detecting %anomalous points in these KPIs.
%This is a challenging problem since anomalies %can take very different forms. 
%For example, it can be a spike, a dip, a %sustained burst, or a complete change in the %pattern of the KPI. 
% %
% Figure \ref{fig:KPI_example} gives some examples of them. 
% \begin{figure}[!t]
%     \centering
%     \includegraphics[scale = 0.5]{figures/KPI_example.png}
%     \caption{Example of an annotated KPI, red points indicate abnormal measures, green point represents a change of concept (concept drift).}
%     \label{fig:KPI_example}
% \end{figure}
%
To date, many anomaly detection algorithms have been proposed, including both supervised and unsupervised ones. 
It is commonly recognized that the supervised anomaly detection algorithms perform better than the unsupervised ones if the labels are sufficiently provided. 
However, this assumption does not always hold true considering the fact that labeling tons of KPIs is extremely time- and effort-consuming and error-prone. 
Unsupervised ones are therefore preferred over the supervised ones in practice. 
One question naturally arising here is that, can we balance the trade-off between labeling effort and detection performance?
In other words, human operators only pay a reasonable amount of effort to label the KPIs of interest (e.g., the ones lead to false positives), and then guide the algorithm towards a better detection behavior. 
This is particularly interesting considering the fact that unsupervised anomaly detection algorithms often suffer from excessive false alarms, especially in the presence of concept drifts potentially resulting from network routine updates or legitimate changes. 

In fact, the aforementioned question has found some answers in the community~\cite{das_active_2019,wang2020practical}, which share the relevant theoretical foundation with active learning. 
With the same question in mind, in this paper, we intend to present a new active learning based solution to improve the performance of an unsupervised anomaly detection algorithm.
% (i.e., Isolation Forest).   
%we propose a feedback-guided anomaly detection %approach. Starting with an unsupervised anomaly %detection model, we use active learning to %intelligently query interesting points from %an expert (oracle) and then update them odel in %order to make it better. 
%
%However, a near complete lack of good quality %labeled datasets is very common in practical %situations, discouraging the use of supervised %learning approaches.
%
%Unsupervised learning methods do not need %labels, but often have to fine-tune some of %their hyperparameters in order to have good %performance. 
%This in turn requires some labels or access to %domain knowledge, which might not always be %available. 
% 
%Feedback-guided methods seem promising not only %because it can address the label problem of %supervised methods, but also the fine-tuning %problem of unsupervised ones. 
%They are also suitable for change-of-concept %situations in which the behavior of KPIs %changes due to, for example, a system update.
%
Specifically, our contributions are four-fold: (1) we develop three query strategies to obtain the most informative and representative positive samples for labeling; (2) we propose a light-weight model update strategy to efficiently derive more accurate anomaly scores and optimal decision threshold, solving the parameterization issue of unsupervised learning methods, and eventually contributing to the improved detection performance; (3) the proposed methods are systematically integrated into an unsupervised anomaly detection algorithm (Isolation Forest), clearly illustrating a feasible approach to introducing human operator's feedback into the closed-loop pipeline for improving its adaptability and performance; (4) a set of experiments with two different datasets are carried out for comparative studies with the state-of-the-art approaches, demonstrating the strong generalization capability of detecting various anomalies in different time series dataset.

The remainder of this paper is organized as follows. 
We firstly review the related work in Section \ref{sec:related_work} and introduce the unsupervised solution to time series anomaly detection problem in Section \ref{sec:unsupervised_solution}. Section \ref{sec:al_solution} describes in detail our active learning solution with experimental results in Section \ref{sec:evaluation}. 
Section \ref{sec:conclusion} finally concludes the paper.

\section{Related Work}
\label{sec:related_work}

\paragraph{Supervised learning.} 
Statistical models such as ARIMA \cite{yu2016improved} have been traditionally used to model the normal behaviors of time series by training on ``clean" data. The deviation from the model forecast is then used as a measure of abnormality.
However, directly thresholding this deviation is usually insufficient in real-world applications. Liu et al.\ \cite{liu2015opprentice} used statistical models for feature extraction and then a classifier such as Random Forest (RF) \cite{Breiman01} to detect the anomalies.
More recently, deep learning algorithms, such as LSTMs \cite{pang2021deep,malhotra2015long}, have been introduced to work as feature extractors in supervised anomaly detection.

\paragraph{Unsupervised learning.} 
There has been a growing interest in unsupervised methods for time series anomaly detection in order to overcome the lack of labeled data in real world scenarios.
For example, Luminol \cite{linkedin}, developed by LinkedIn, segments time-series into chunks and uses the frequency of similar chunks to calculate the anomaly scores. 
SPOT and DSPOT \cite{siffer2017anomaly} use extreme value theory to model distribution tail in order to detect outliers in time series. Microsoft \cite{ren_time-series_2019} uses spectral residual (SR) concept from signal processing to develop their SR-based anomaly detector.
More recently, deep learning-based methods \cite{pang2021deep} have been also employed to detect anomalies in unsupervised settings. For example, DONUT \cite{xu2018unsupervised} uses variational auto-encoder to detect anomalies from seasonal KPIs.

\paragraph{Active learning.}
There are two major active learning approaches for anomaly detection.
The first approach, such as the one proposed in \cite{gornitz2013toward}, usually solves a semi-supervised learning problem (SSAD) that uses both labeled and unlabeled points in its underlying formulation. 
When no labels are available, the models are first trained on unlabeled data in unsupervised settings. They are subsequently updated by incorporating labeled points from feedback into the learning problem. In \cite{trittenbach2020overview}, the authors used variants of SSAD model for benchmarking and found that there is no one-fit-all strategy for one-class active learning. 
Recently, Amazon developed NCAD based on deep semi-supervised learning \cite{ruff2020deep} for time series anomaly detection \cite{carmona2021neural}.

The second approach to active anomaly detection is based on ensemble learning. 
The base learners are usually tree-based, such as Isolation Forest (iForest) \cite{liu2008isolation}, RS-Forest \cite{wu2014rs}, or Robust Random Cut Forest (RRCF) \cite{guha2016robust}. They are firstly trained on unlabeled data and then be updated using labeled points from the feedback. 
The update can be in the form of adjusting the weights of trees \cite{wang2020practical}, weights of trees' nodes \cite{das_active_2019} or trees' edges \cite{siddiqui_feedback-guided_2018} in order to improve the performance of the base models on these labeled points.

For sample selection, the existing methods explored three strategies. 
\emph{Top anomalies} implies the selection of points that have the maximum anomaly scores. \textit{Top diverse} strategy, which is similar to the previous one, but requires the maximization of certain ``diversity" measure in the group of selected points. 
\textit{Random} strategy selects points randomly and is usually used as a baseline.

%{\color{red} Attention, you should review literature in 2 steps. The first one is supervised + unsupervised (because lack of data). And the second is on feedback-based methods in order to benefit from a small set of labeled data.}

%{\color{blue} At the end of each step, you should provide some discussions of their strengths and limitations.}

%Now may be talk about the fact that these have some limitations

%- Common challenges of Anomaly detection (make report version short and compact)
%- Parameterization : To make verbose \\
%As a solution AL, but this one has also some challenges
%- How to choose samples \\
%- How to update the model \\

%Due to the lack of good quality labeled data, there has been a huge interest (especially from technology industries) for discovering unsupervised models that can detect anomalies without requiring labels.

\section{Unsupervised Anomaly Detection}
\label{sec:unsupervised_solution}

Let $\mathbf{X} = (x_1, x_2, .., x_n)$ be a time series, which is a sequence indexed in time with $x_i \in \mathbb{R}^d$. 
In this work, we focus on univariate time series data where $d=1$. 
A time series anomaly detector usually takes $\mathbf{X}$ as input and outputs a sequence $\mathbf{Y} = (y_1, y_2, ..., y_n)$ of the same length, where $y_i = 1$ if $x_i$ is anomalous or $y_i = 0$ otherwise.
An anomaly detection pipeline is usually composed of two main modules: feature extraction and anomaly detection model.

\paragraph{Features extraction.}
A feature extractor projects the input sequence $\mathbf{X} = (x_1, x_2, .., x_n)$ into a feature space of dimension $d_f$ so that it becomes easier to distinguish anomalous points from normal ones. Each point $x_i \in \mathbf{X}$ is then represented by a vector $x^{\prime}_i$ of size $d_f$.
In this work, we extract features in online mode and uses sliding windows of size $w=5$. 
More specifically, for a given timestamp $t$, we calculate some measures from the window $[x_{t-w},...,x_{t-1}]$ and subtract them from the current value $x_t$.
Among $d_f=6$ features used in this work and listed in Table \ref{table:feature_definition}, five of them are statistical features widely used in time series anomaly detection. The last one is the saliency map calculated from one-day-length subsequence \cite{ren_time-series_2019}.
\begin{table}[H]
    \vspace{-0.5cm}
    \small
    \centering
    \caption{The 6 features used to represent each point in the feature space.} 
    \begin{tabular}{|l|l|}
        \hline
        \multicolumn{1}{|c|}{\textbf{Feature}}
        & \multicolumn{1}{c|}{\textbf{Description}}
        \\ \hline
        \texttt{max}
        & Difference with maximum value on window of size $w$
        \\ \hline
        \texttt{min}
        & Difference with minimum value on window of size $w$
        \\ \hline
        \texttt{mean}
        & Difference with mean of window of size $w$
        \\ \hline
        \texttt{naive}
        & Difference from the previous value
        \\ \hline
        \texttt{linear\_residual}
        & Fit a linear model on a window of size $w$ 
        \\[-2pt]
        & and compute the residual at the current point
        \\ \hline
        \texttt{saliency\_map} & Spectral saliency at the current point
        \\ \hline
    \end{tabular}
    \label{table:feature_definition}
\end{table}
\paragraph{Anomaly detection model.}
Similar to \cite{das_active_2019}, we use iForest as the anomaly detection model because it is an unsupervised model being composed of trees and is much faster than RRCF.  
The initial iForest model is trained on the pool of unlabeled points. 
An ensemble of trees also facilitates the model update in active learning. 

Ensemble learning methods, especially the tree-based ones such as iForest, are well suited for active learning anomaly detection. 
This is because anomalies usually exhibit abnormal behaviors and their representations are mostly scattered in the feature space, which is in contrast to normal points whose representations form high density clusters. 
The separation boundaries between the representation of normal and abnormal points in the feature space are thus \emph{non-homogeneous}, which can be well-represented by a properly trained iForest model.
In addition, since iForest is widely used in outlier and anomaly detection, having a mechanism to enhance it using active learning would benefit the whole community.

\section{Active Anomaly Detection}
\label{sec:al_solution}

\subsection{Design assumptions}
\label{subsec:al_assumption}
\begin{figure}[!t]
    \centering
    \includegraphics[scale = 0.7]{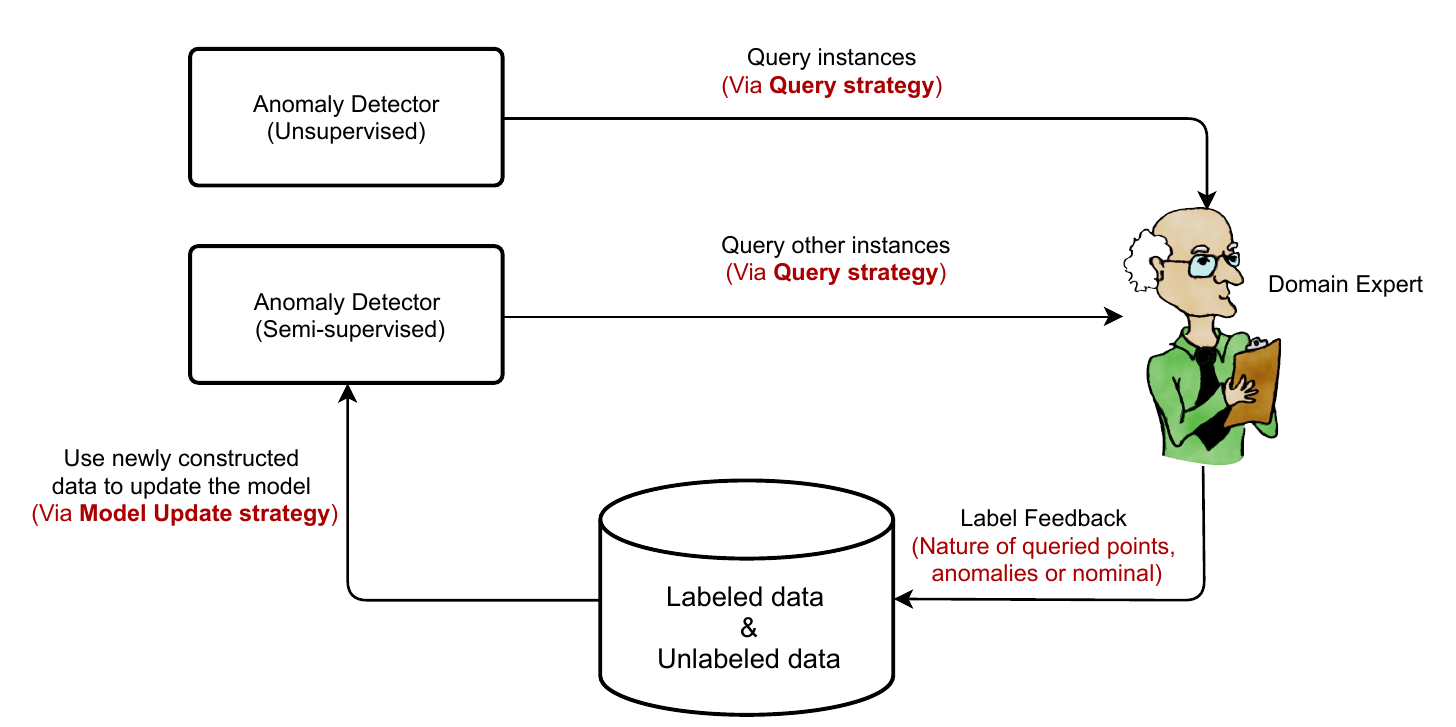}
    \caption{Design workflow: use active learning to improve unsupervised anomaly detection}
    \label{fig:AL_pipeline}
\end{figure}

While there are different variants of active learning scenarios in the literature,
our focus in this paper is on the \emph{pool-based active learning} \cite{hanneke2014theory} only. Specifically, we assume that a (typically large) number of unlabeled data points, referred to as the \emph{unlabeled pool}, is available and accessible to the learning process.  
An analyst or a domain expert is also available to provide a ground-truth label for any point in this pool upon request.  The requests to domain expert are in ``batch", consisting of all ``interesting" points within the given budget. 
Figure \ref{fig:AL_pipeline} illustrates the active learning process that has three main components, (1) start with a fully unsupervised model; 
(2) select points according to the given budget and query strategy; 
(3) update the model based on domain expert's feedback. 
This process can be iterative, either upon a request, or is automatically triggered when the model performance gets worse than a certain threshold.  
%as long as there is still a need to update the %model or the labeling budget is increased.
%

Our active learning pipeline is essentially inspired from  \cite{das_active_2019} and \cite{wang2020practical}. They both use tree-based ensemble unsupervised models as the anomaly detectors. Wang et al. \cite{wang2020practical} handles the ensemble at the tree level, and the model update focuses on adjusting the weights of trees in the ensemble. Das et al. \cite{das_active_2019}, on the other hand, works at the nodes of constituting trees and updates the model by adjusting node weights. These two approaches, however, have high complexity. 
\begin{itemize}
    \item  Das et al. \cite{das_active_2019} uses iForest as the base model, which is relatively fast. However, its model updating process is computationally expensive since node-level features and scores for each point need to be computed, either in training or inference. In addition, it solves an underlying optimization problem in order to find the best node weights (\textbf{NW}), which can be too expensive for a model with a high number of nodes. 
    \item Wang et al. \cite{wang2020practical} employs a very fast and straightforward model update strategy which consists of adjusting tree weights according to the scores of anomalous points. It, however, uses a slightly modified version of RRCF as the base model. It's well known that RRCF is slow since it is designed for streaming context and the model auto-adjusts for each incoming stream value in order to adapt to the new data distribution.
    % (or at least, its open sourced implementation, the reason behind is explained in \cite{klabum}). 
\end{itemize}

In order to build a very fast and easy-to-deploy anomaly detector with active learning, we propose an active learning pipeline using iForest as the base model. iForest model is updated by adjusting the weights of its constituting and by seeking the best value for its ``offset" parameter. The remaining of this section will describe in details different query and model update strategies we adopt in this work.

\subsection{Query strategy}
\label{subsec:query_strategy}
There exists a number of query methods in the literature, and they usually follow a common formulation.
Given $\mathcal{U}$ the set of unlabeled points, $b$ the given budget, and an interest function $f: \mathcal{U} \to \mathbb{R}$ used as a measure of ``informativeness" or ``utility" of requesting the label for each point $x \in \mathcal{U}$, a query strategy aims at selecting $x_{i=1,\ldots ,b} \in \mathcal{U}$ in order to maximize $\sum_{i=1}^b f(x_i)$.
The choice of a query strategy thus usually reduces to the choice of an interest function $f$. In this work, we use three query strategies for our active learning pipeline.
Algorithm \ref{alg:query_strategy} presents how points are selected according to $\mathcal{U}$, $b$, and $f$.
\begin{center}
% \begin{small}
\begin{minipage}{.82\linewidth}
\vspace{-0.75cm}
\begin{algorithm}[H]
	\caption{\texttt{select\_points}($\mathcal{U}$, $b$, $f$)}
	\label{alg:query_strategy}
	\begin{algorithmic}
		\STATE \textbf{Input:} $\mathcal{U}$ (unlabeled dataset), $b$ (budget), $f$ (interest function)
		\STATE Set ${\bf Q} \leftarrow \emptyset$
		\WHILE{$|{\bf Q}| < b$}
		\STATE Let ${\bf x} \leftarrow \underset{x \in \mathcal{U} \setminus {\bf Q}}{\operatorname{argmax}} $ $f(x) $
		\STATE Set ${\bf Q} \leftarrow {\bf Q} \cup \{{\bf x}\}$ 
		\ENDWHILE
		\RETURN ${\bf Q}$
	\end{algorithmic}
\end{algorithm}
\end{minipage}
% \end{small}
\end{center}

\begin{figure}[!t]
    \centering
    \includegraphics[scale = 0.25]{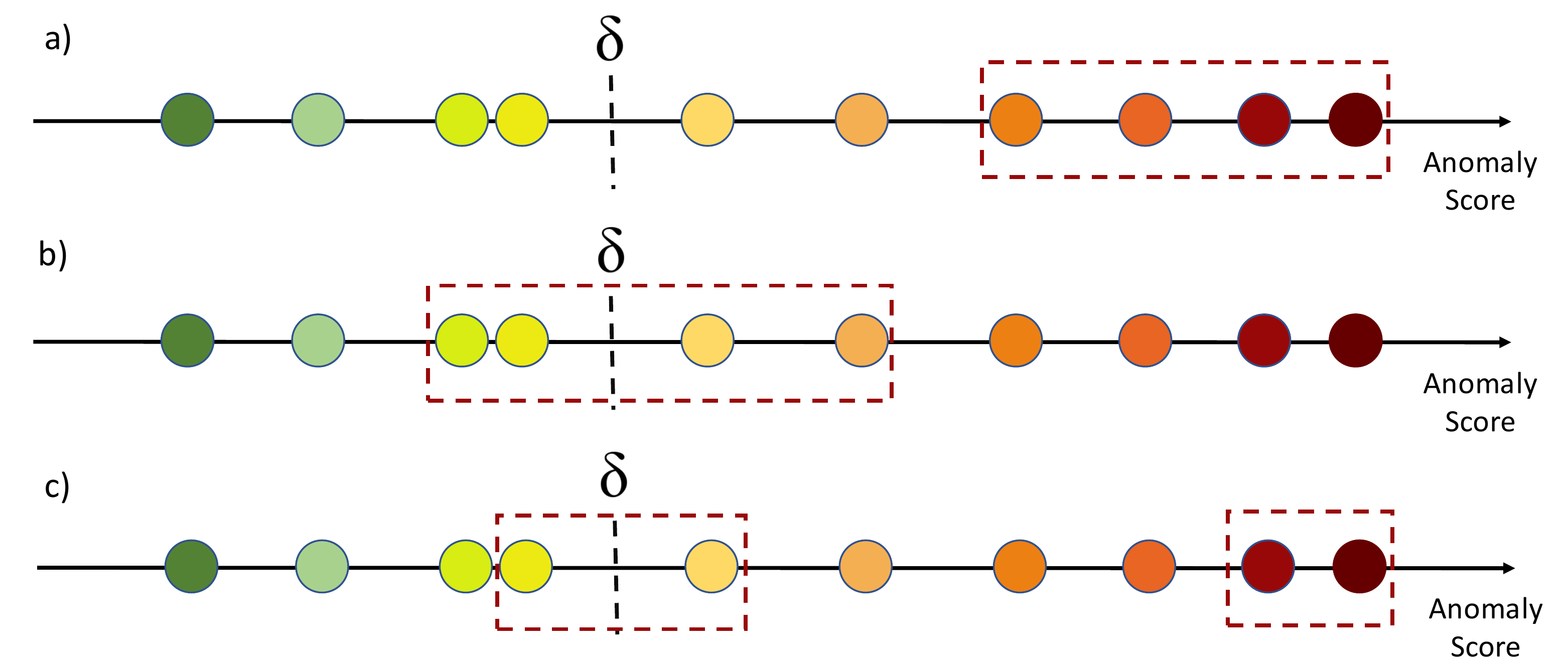}
    \caption{\small{Illustration of query strategies using a dataset of 10 points sorted according to their anomaly scores (higher scores $\rightarrow$ right). With a budget $b=4$, points inside the rectangle are selected according to a) TA, b) CTDB and c) TA + CTDB strategies.}}
    \label{fig:all_strategies}
\end{figure}

\begin{itemize}
    \item Top anomalous selection \textbf{(TA)} (Figure \ref{fig:all_strategies}(a)): Also called ``greedy strategy" in the context of anomaly detection, it selects points that have the highest anomaly scores. 
    Let $\mathcal{UAD}(x)$ be the anomaly score of $x$, the interest function corresponding is defined as $f(x) = \mathcal{UAD}(x)$. 
    \item Close to decision boundary selection \textbf{(CTDB)} (Figure \ref{fig:all_strategies}(b)): Points that are the closest to the decision boundary of the anomaly detector are selected. The region near the decision boundary is expected to contain the most difficult points to classify.
    Mathematically, if we denote $\delta$ the threshold used to classify points, the interest function is defined as $f(x) = -( \mathcal{UAD}(x) - \delta)^2$.
    \item \textbf{TA + CTDB} (Figure \ref{fig:all_strategies} (c)): This is a combination of the two above strategies by using half of the budget for \textbf{TA} and the other half for \textbf{CTDB}. This combination gives more diversity in the selected samples and is expected to better update the unsupervised model.
\end{itemize}

\subsection{Model update strategy}
\label{subsec:model_update_strategy}
For model update, we rely on the anomaly scores obtained by the iForest model. After querying a set of points for their labels, the unsupervised iForest model can be updated using one of the following strategies:
\begin{itemize}
    \item Tree weights update \textbf{(TW)}: iForest is a tree-based model and, in the original formulation of its scoring function, its trees contribute equally to the calculation of anomaly score for each input point. 
    We adopt the strategy proposed in \cite{wang2020practical} to adjust the contributions or weights of iForest trees so that if a tree turns out to be more accurate in its anomaly scoring of anomalous queried points, it contributes more to the calculation of anomaly score. 
%     %

    \item Offset update \textbf{(O)}: iForest has a critical hyperparameter called \emph{contamination ratio}. It is used during training to determine, thanks to the percentile metric,
the offset value, which will be used in inference to classify points as anomalous
or nominal by thresholding their anomaly scores.
    In practice, this ratio is usually guessed based on application context, and it usually turns out to be very difficult to set it properly. We propose to ``learn" this offset value based on the feedback for queried points.
    Finding the best offset from feedback can be done in various ways, such as a simple rule-based thresholding or training a classifier and then using its decision boundary as the learned offset. 
    We have tested and found that the rule-based thresholding has performed better than a linear SVM. Algorithm \ref{alg:update_model_offset} presents the procedure to calculate offset value from feedback.
\begin{center}
\vspace{-0.5cm}
\begin{minipage}{.9\linewidth}   
    \begin{algorithm}[H]
	\caption{\texttt{calculate\_offset}(${\mathcal L}$, $\mathbf{S}$)}
	\label{alg:update_model_offset}
	\begin{algorithmic}
		\STATE \textbf{Input:} $\mathcal{L}$ (dataset of $n$ labeled points) and $\mathbf{S}$ (anomaly scores of points in $\mathcal{L}$ by iForest model)
		\STATE Set ${\mathcal L}_a \leftarrow \{x \in \mathcal L : x \text{ is labelled anomalous} \}$
		\STATE Set ${\mathcal L}_n \leftarrow \{x \in \mathcal L : x \text{ is labelled normal} \}$
        \STATE $\text{offset} = (\min\left(\{\mathbf S(x): x \in {\mathcal L}_a \}\right) + \max\left(\{\mathbf S(x): x \in {\mathcal L}_n \}\right)) / 2$
		\RETURN offset
	\end{algorithmic}
    \end{algorithm}
\end{minipage}
\end{center}
    \item \textbf{(TW+O)}: This is a combination of the two aforementioned strategies by applying \textbf{TW} and \textbf{O} in sequence.
\end{itemize}

\begin{figure}[!t]
    % \begin{adjustwidth}{-0.1cm}{-1cm}
    \centering
    %\includegraphics[scale = 0.65]{figures/AL_Iforest.png}
    %\def\svgscale{0.9}
    %\includesvg[width = \textwidth]{figures/distro.svg}
    \includegraphics[width=0.95\textwidth]{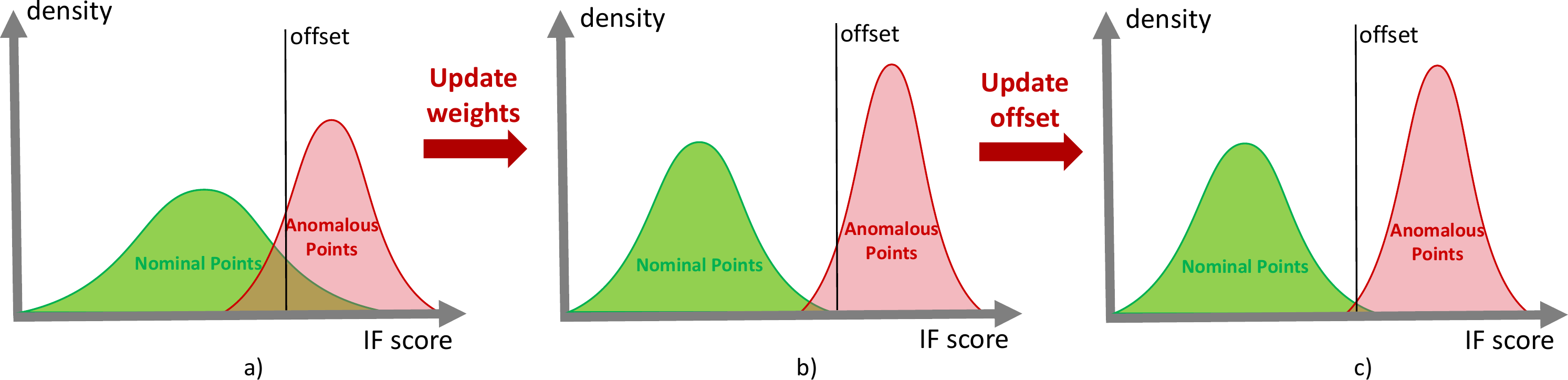}
    % \end{adjustwidth}
    \vspace{-0.5cm}
    \caption{Illustration of the evolution of score distribution and offset of an iForest model under tree weights and offset update strategies.} 
    \label{fig:model_update_illustration}
\end{figure}

\paragraph{Intuition behind the two-step model update strategy:}
\label{subsec:model_update_intuition}
Figure \ref{fig:model_update_illustration} illustrates the impact of the above-stated model update strategies on the score distribution and offset of an iForest model. 
At the beginning, the score distributions of nominal and anomalous points produced by the unsupervised iForest model have a ``large" overlap and the offset value is not properly set to ``well" separate these two distributions (Figure \ref{fig:model_update_illustration}(a)).
By updating the weights of iForest trees,
%using Algorithm \ref{alg:update_trees_weights}, 
these two distributions are pushed further away from each other, causing anomalous and nominal samples to have higher and lower scores, respectively (Figure \ref{fig:model_update_illustration}(b)).
It should be noted that updating the weights of iForest trees has no impact on the offset of the iForest model. This offset value is further adjusted using Algorithm \ref{alg:update_model_offset} so that it can better separate the two score distributions and, consequently, the adjusted iForest model can better distinguish anomalous points from nominal ones (Figure \ref{fig:model_update_illustration}(c)).
Experimental evidence for the impact of each model updating strategy is given in Section \ref{sec:evaluation_active_learning}.
\section{Performance Evaluation}
\label{sec:evaluation}

\subsection{Experimental settings}
\begin{figure}[!t]
    \centering
    \includegraphics[scale=0.72]{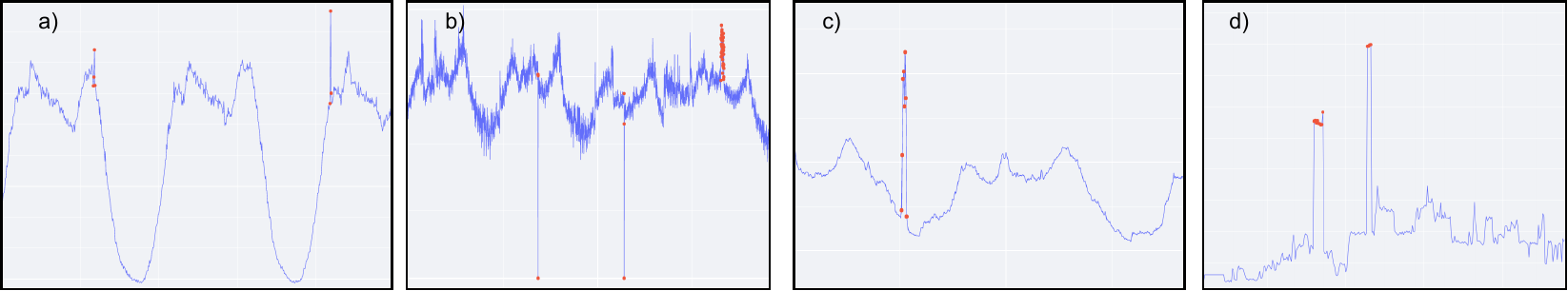}
    \caption{Some KPIs from \textsc{AIOps} ((a) and (b)) and \textsc{Huawei} ((c) and (d)) datasets. Anomalous points are marked using red color.}
    \label{fig:examples_kpis}
\end{figure}

\paragraph{Datasets:}
Our evaluation experiments are conducted on two datasets. The first one (\textsc{AIOps}) is public and is commonly used for performance evaluation of univariate time series anomaly detection algorithms. It is released by the AIOps2018 competition\footnote{https://github.com/NetManAIOps/KPI-Anomaly-Detection} and consists of $29$ KPIs collected from some internet companies in China. In our experiments, we use the training and testing data from the \emph{Final} subset of this dataset
The second dataset (\textsc{Huawei}) is private and collected from Huawei production environment. It is composed of $8$ univariate telecom core network KPIs from different network elements. Each KPI is split into two halves, the first for training and the second for testing.
These two datasets contain KPIs with a wide range of time series characteristics and anomalous patterns. Anomalous points and segments are labeled by domain experts and annotated as positive points, whereas nominal ones are designated as negative points.
Some KPIs from these two datasets are shown in Figure \ref{fig:examples_kpis}. Table \ref{table:dataset_statistics} summarizes some statistics of the two datasets.
\begin{table}[!t]
\small
\centering
\caption{Statistics of benchmark datasets}
\begin{tabular}{l|cccc}
\hline
\textbf{Dataset} & \textbf{\#KPIs} & \textbf{\#Points} & \textbf{\#Anomalous points} & \textbf{Sampling interval} \\ 
\hline
\textbf{AIOps}   & 29     & 5922913    & 134114 (2.26\%)  & 1 or 5 minute  \\
% \textbf{Yahoo}   & 367    & 572966     & 3896 (0.68\%)    & 1, 5 minute    \\
\textbf{Huawei}  & 8      & 119744     & 1188 (0.99\%)    & 5 minute       \\ 
\hline
\end{tabular}
\label{table:dataset_statistics}
\end{table}

\paragraph{Metrics:}
For performance evaluation and comparison purpose, we adopt the evaluation protocol suggested by \cite{DBLP:conf/infocom/ZhaoZLLZP19} and commonly used by the community. In this protocol, the F1-score is not calculated directly based on point-wise matching between the labels and detection results. A delay parameter $k$ is introduced to adjust the detection results before F1-score calculation. 
According to human experts and for a contiguous anomaly segment, it is acceptable if the algorithm can detect and trigger an alert within a delay of $k$ points. More precisely, we mark a segment of continuous anomalies as correctly detected if a point from this segment that is within $k$ points from the segment's beginning is detected. In our experiments, the delay for both \textsc{AIOps} and \textsc{Huawei} datasets is $k = 7$, as recommended by the AIOps competition and used in other works. % %
% Figure \ref{fig:evaluation_protocol} illustrates the adjustment for a sample label and detection result.
% %

\subsection{Supervised and unsupervised anomaly detection}
\label{sec:evaluation_baseline}
\vspace{-3pt}
To demonstrate the effectiveness of our anomaly detection pipeline, we compare it with several SoTA anomaly detection methods in Table \ref{table:sup_unsup_results}. In addition to Isolation Forest (iForest), we also use Random Forest (RF) and run our pipeline presented in Section \ref{sec:unsupervised_solution} in a supervised setting in order to establish the performance upper bound when all points are labeled in our active anomaly detection (i.e., query budget = $100\%$). RF is selected because it is also a tree-based ensemble model.

It can be seen that our best RF and iForest models achieve the best results among all supervised (sup.) and unsupervised (un.) methods, respectively. Thus, with a relatively simple and right set of features, combined with the right parameterization of popular supervised and unsupervised models, we can achieve the SoTA performance on benchmark datasets.
This observation is valuable in a practical viewpoint, in which simple and explainable features and models are usually preferable over more complex ones.
Also, the performance of iForest drops significantly when it uses, for example, an inappropriate value for its \emph{contamination} parameter, among others. We assume that this type of performance degradation due to parameterization also applies to all other methods. 
This issue, however, can be handled by active learning, as will be shown in the next section.
\begin{table}[t]
    \small
    \centering
    \caption{Performance comparison with SoTA methods. Random Forest and iForest models are trained with features in Table \ref{table:feature_definition}. For iForest, the value in bracket is the contamination ratio: $0.01$ (usually recommended), $0.03$, and $0.007$ (best value found by grid search on training data). }
    \label{table:sup_unsup_results}
    \begin{tabular}{l|c|cc@{\hskip 0.15in}||@{\hskip 0.15in}l|c|ccccccc}
        \toprule
        Model   & Type  & \textsc{AIOps} & \textsc{Huawei} & 
        Model   & Type  & \textsc{AIOps} & \textsc{Huawei} \\ 
        \midrule
        SPOT        \cite{siffer2017anomaly}    & un.  & 21.7 & --- &
        NCAD (un.)  \cite{carmona2021neural}    & un.  & 76.6 & --- \\
        DSPOT       \cite{siffer2017anomaly}    & un.  & 52.1 & --- &
        NCAD (sup.) \cite{carmona2021neural}    & sup. & 79.2 & --- \\
        DONUT       \cite{xu2018unsupervised}   & un.  & 72.0 & --- & 
        RF (best)                               & sup. & \textbf{81.2} & \textbf{72.6} \\
        SR          \cite{ren_time-series_2019} & un.  & 62.2 & 40.5 &
        iForest (0.01)                          & un.  & 73.3 & 51.8 \\
        SR-CNN      \cite{ren_time-series_2019} & un.  & 77.1 & --- &
        iForest (0.03)                          & un.  & 51.3 & 40.8 \\
        SR-DNN      \cite{ren_time-series_2019} & sup. & 81.1 & --- &
        iForest (best)                    & un.  & \textbf{78.4} & \textbf{65.4} \\
        \bottomrule
    \end{tabular}

\end{table}

It should be noted that NCAD \cite{carmona2021neural} does not strictly follow the adopted evaluation protocol. It uses a more relaxed one without a delay restriction, which is thus equivalent to the adopted protocol with $k=\infty$.
SR-CNN is a semi-supervised model that requires 65 million anomaly-free simulated points to train its CNN model for saliency map thresholding \cite{ren_time-series_2019}. 
Finally, the gap in performance between RF and a ``perfect" model can be explained partially by the lack of coherence in dataset labeling and by the limited representative and expressive power of the feature set and model.

\subsection{Active anomaly detection}
\label{sec:evaluation_active_learning}
We evaluate our active anomaly detection pipeline on the two benchmark datasets and provide the results in Table \ref{table:al_results}. 
We are able to compare our approach with \cite{das_active_2019} using its open source implementation\footnote{https://github.com/shubhomoydas/ad\_examples}. Since the implementation of  \cite{wang2020practical} is not open, we've implemented it using an open source implementation of RRCF \cite{bartos_2019_rrcf}. 
Our implementation turns out to be too slow due to the high complexity of RRCF. We couldn't obtain experimental results in a reasonable amount of time, and thus decided to not compare our method with \cite{wang2020practical} here.

In all experiments, iForest (0.03) presented in Section \ref{sec:evaluation_baseline} is used as the baseline unsupervised model. Each value in Table \ref{table:al_results} represents the F1-score obtained by using a unique combination of query strategy, model update strategy, and query budget.
It can be seen that, even starting with a weak unsupervised model and using only $1\%$ query budget, our active learning pipeline improves the performance by $56.51\%$, reaching $80.29\%$ F1-score for \textsc{AIOps} dataset. Similarly, for \textsc{Huawei} dataset, the performance is improved by $75.14\%$ and reaches $71.37\%$ F1-score. This performance is better than the performance achieved by \cite{das_active_2019} and iForest (best) and is very close to the performance of RF (best) shown in Table \ref{table:sup_unsup_results}. This clearly demonstrates the effectiveness of our active learning pipeline.

Among query strategies, greedy selection (\textbf{TA}) outperforms random and \textbf{CTDB}. The random strategy does not really improve the performance. This is similar to the observations reported in previous works \cite{das_active_2019,wang2020practical} and demonstrates the necessity of a good query strategy in an active learning pipeline. Among model update strategies, \textbf{TW} has almost no impact, regardless the amount of query budget. \textbf{O} and \textbf{TW+O} lead to the best improvement for \textsc{AIOps} and \textsc{Huawei} datasets, respectively. This demonstrates the importance of our proposed offset update (\textbf{O}) strategy.
Finally, using our active learning approach, the required query budget to reach the best improvement is small compared to \cite{das_active_2019}. Our approach needs about $1\%$ whereas \cite{das_active_2019} needs $25-50\%$ of the training data.

In terms of computational complexity, our approach is about $14\times$ and $6\times$ faster than \cite{das_active_2019} on \textsc{AIOps} and \textsc{Huawei} datasets. This confirms the utility of our simple model update strategies, compared to the more complex ones used in \cite{das_active_2019}. 
%\markred{Maybe we can highlight that this gap can even get greater with number of KPIs (Huawei$<$AIOps)} \markblue{This is because the budget is proportional, $1\%$ of larger dataset give you more points than smaller dataset. The model trained from larger dataset is usually more complex also.}
%
\begin{table}[!t]
\label{table:al_results}
\caption{Experimental results on active anomaly detection. B/L indicate the iForest (0.03) model. 1\%, 5\%, 25\% and 50\% are the allowed query budgets. Each time value indicates the training and inference time to generate all results of the same row (B/L, 1\%, 5\%, 25\% and 50\%) using 10 Intel Xeon CPUs (E5-2690 v3 @ 2.60GHz).}
% \begin{adjustwidth}{-1.3cm}{}

\resizebox{1.\columnwidth}{!}
{
\begin{tabular}{ccc|cllllr|cllllr|}
\cline{4-15}
& & & \multicolumn{6}{c|}{\textsc{AIOps}} 
& \multicolumn{6}{c|}{\textsc{Huawei}}   \\ \cline{2-15} 
\multicolumn{1}{c|}{} 
& \multicolumn{1}{c|}{\begin{tabular}[c]{@{}c@{}}Query \\ strategy\end{tabular}}
& \begin{tabular}[c]{@{}c@{}}Model update\\   Strategy\end{tabular} 
& B/L & \multicolumn{1}{c}{1\%} & \multicolumn{1}{c}{5\%} & \multicolumn{1}{c}{25\%} 
& \multicolumn{1}{c}{50\%} & \multicolumn{1}{c|}{\begin{tabular}[c]{@{}c@{}}time \\ (s)\end{tabular}} 
& B/L & \multicolumn{1}{c}{1\%} & \multicolumn{1}{c}{5\%} & \multicolumn{1}{c}{25\%} 
& \multicolumn{1}{c}{50\%} & \multicolumn{1}{c|}{\begin{tabular}[c]{@{}c@{}}time\\ (s)\end{tabular}} \\
\hline
\multicolumn{1}{|c|}{} & \multicolumn{1}{c|}{TA} &   
& & 48.87 & 53.93 & 53.40     & \textbf{71.82}     & 6964  
&  & 60.21 & 42.55 & 69.22     & \textbf{70.74}     & 208  
\\ \cline{2-2} \cline{5-9} \cline{11-15} 
\multicolumn{1}{|c|}{\multirow{-2}{*}{\textbf{\cite{das_active_2019}}}} 
& \multicolumn{1}{c|}{Random} & \multirow{-2}{*}{NW}     
& & 27.91 & 54.90 & 50.27     & 62.30     & 8133  
&  & 3.08     & 51.66 & 47.49     & 61.36     & 252  \\
\cline{1-3} \cline{5-9} \cline{11-15} 
\multicolumn{1}{|c|}{} & \multicolumn{1}{c|}{}   & TW   
& & 51.77 & 51.55 & 51.42     & 51.44     & 427   
&  & 60.73 & 60.41 & 60.17     & 60.17     & 29.3 \\
\multicolumn{1}{|c|}{} & \multicolumn{1}{c|}{}   & O 
& & \textbf{80.29}    & 75.02 & 60.37     & 53.82     & 533   
&  & 70.86 & 71.34 & 57.55     & 53.10     & 32.3 \\
\multicolumn{1}{|c|}{} & \multicolumn{1}{c|}{\multirow{-3}{*}{TA}}      & TW+O  
& & 80.24 & 74.88 & 60.18     & 53.67     & 520   
&  & \textbf{71.37}    & 71.29 & 55.83     & 52.47     & 33.8 \\ 
\cline{2-3} \cline{5-9} \cline{11-15} 
\multicolumn{1}{|c|}{} & \multicolumn{1}{c|}{}   & TW   
& & 50.93 & 50.97 & 51.22     & 51.34     & 453   
&  & 59.31 & 60.17 & 60.17     & 60.17     & 29.2 \\
\multicolumn{1}{|c|}{} & \multicolumn{1}{c|}{}   & O 
& & 52.10 & 52.78 & 53.68     & 52.01     & 516   
&  & 59.49 & 60.10 & 57.55     & 53.10     & 32.5 \\
\multicolumn{1}{|c|}{} & \multicolumn{1}{c|}{\multirow{-3}{*}{CTDB}}    & TW+O  
& & 52.69 & 53.20 & 53.49     & 51.75     & 500   
&  & 60.47 & 61.18 & 55.83     & 52.47     & 33.8 \\
\cline{2-3} \cline{5-9} \cline{11-15} 
\multicolumn{1}{|c|}{} & \multicolumn{1}{c|}{}   & TW   
& & 51.76 & 51.55 & 51.34     & 51.35     & 509   
&  & 60.17 & 60.41 & 60.10     & 60.17     & 33.2 \\
\multicolumn{1}{|c|}{} & \multicolumn{1}{c|}{}   & O 
& & 77.03 & 75.47 & 67.16     & 60.32     & 610   
& & 47.53 & 70.90 & 63.32     & 57.55     & 37.4 \\
\multicolumn{1}{|c|}{\multirow{-9}{*}{\textbf{Ours}}} 
& \multicolumn{1}{c|}{\multirow{-3}{*}{\begin{tabular}[c]{@{}c@{}}TA +\\ CTDB\end{tabular}}} & TW+O  
& \multirow{-11}{*}{51.30} & 52.19 & 52.24 & 53.04     & 53.64     & 610   
& \multirow{-11}{*}{40.75} & 60.79 & 60.91 & 61.91     & 55.83     & 38.3 
\\ \hline
\end{tabular}
}
% \end{adjustwidth}
\end{table}

\begin{samepage}
\section{Concluding Remarks}
\label{sec:conclusion}

This paper proposed an efficient active learning based approach to systematically integrating the feedback and expert knowledge of network operators into unsupervised anomaly detection pipeline. 
In particular, we proposed three effective query strategies to assist operator in labeling those KPI samples that lead to alarms, including both true and false ones.
A lightweight model update algorithm, which consists of updating the weights of trees and the adjustment of decision threshold in Isolation Forest model, has been also developed to improve the performance and efficiency.
The experiments with two datasets have validated the performance advantages over baseline Isolation Forest and existing active learning based method in terms of detection performance and computational efficiency. 
Despite the claimed advantages, we believe sample query strategy and model update algorithm, as well as their integration with other unsupervised anomaly detection algorithms (e.g., RRCF), still have room to be further improved.  
\end{samepage}

%
% ---- Bibliography ----
%

\bibliographystyle{splncs04}
\bibliography{bibliography}

\end{document}